\documentclass[runningheads]{llncs}

\pdfoutput=1

\usepackage{graphicx}

\usepackage[cmex10]{amsmath} 
\usepackage{amssymb}
\usepackage{mathtools}

\usepackage{graphicx}

\usepackage{tikz}
\usetikzlibrary{arrows}
\usepackage{verbatim}

\usepackage{multirow}

\usepackage{xcolor}
\usepackage{hyperref}

\begin{document}

\title{Vision Transformer for Fast and Efficient Scene Text Recognition}
%
%
\author{Rowel Atienza\orcidID{0000-0002-8830-2534} } 
%
%
\institute{Electrical and Electronics Engineering Institute\\University of the Philippines\\\email{rowel@eee.upd.edu.ph}} 
%
\maketitle              
\begin{abstract}
Scene text recognition (STR) enables computers to read text in natural scenes such as object labels, road signs and instructions. STR helps machines perform informed decisions such as what object to pick, which direction to go, and what is the next step of action. In the body of work on STR, the focus has always been on recognition accuracy. There is little emphasis placed on speed and computational efficiency which are equally important especially for energy-constrained mobile machines. In this paper we propose ViTSTR, an STR with a simple single stage model architecture built on a compute and parameter efficient vision transformer (ViT). On a comparable strong baseline method such as TRBA with accuracy of 84.3\%, our small ViTSTR achieves a competitive accuracy of 82.6\% (84.2\% with data augmentation) at $2.4\times$ speed up, using only 43.4\% of the number of parameters and 42.2\% FLOPS. The tiny version of ViTSTR achieves 80.3\% accuracy (82.1\% with data augmentation), at $2.5\times$ the speed, requiring only 10.9\% of the number of parameters and 11.9\% FLOPS. With data augmentation, our base ViTSTR outperforms TRBA at 85.2\% accuracy (83.7\% without augmentation) at $2.3\times$ the speed but requires 73.2\% more parameters and 61.5\% more FLOPS.  In terms of trade-offs, nearly all ViTSTR configurations are at or near the frontiers to maximize accuracy, speed and computational efficiency all at the same time. 

\end{abstract}

\begin{figure}
\centering

    \includegraphics[scale=0.46]{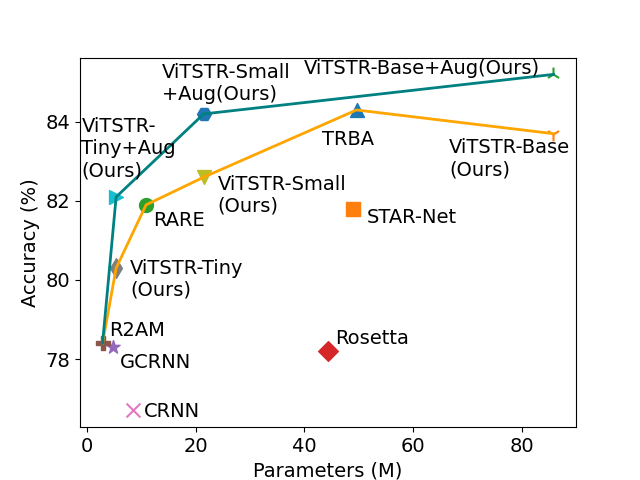}
        
    \includegraphics[scale=0.46]{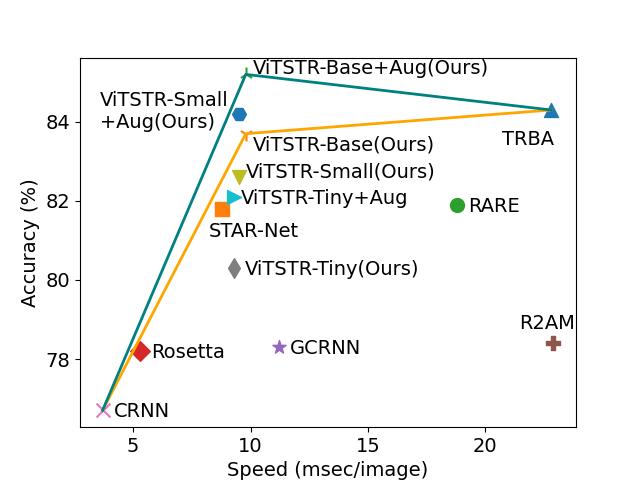}
    
    \includegraphics[scale=0.46]{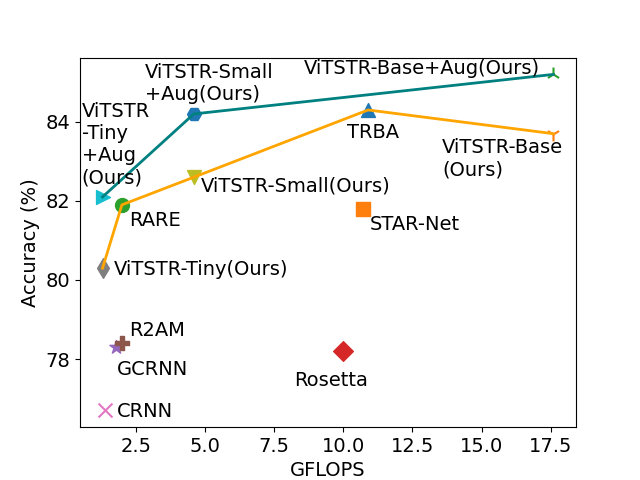}
 \caption{Trade-offs between accuracy vs number of parameters, speed and computational load (FLOPS). +Aug uses data augmentation. Almost all versions of ViTSTR are at or near the frontiers to maximize the performance on all metrics. The slope of the line is the accuracy gain as the number of parameters, speed or FLOPS increases. The steeper the slope, the better. Teal line includes ViTSTR with data augmentation.}
\label{fig:acc_vs_perf}
\end{figure}

\section{Introduction}
STR plays a vital role for machines to understand the human environment. We invented text to convey information through labels, signs, instructions and announcements. Therefore, for a computer to take advantage of this visual cue, it must also understand text in natural scenes. For instance, a "Push" signage on a door tells a robot to push it to open. In the kitchen, a label with "Sugar" means that the container has sugar in it. A wearable system that can read "50" or "FIFTY" on a paper bill can greatly enhance the lives of visually impaired people.

STR is related but different from the more developed field of Optical Character Recognition (OCR). In OCR, symbols on a printed front facing document are detected and recognized. In a way, OCR operates in a more structured setting. Meanwhile, the objective of STR is to recognize symbols in varied unconstrained settings such as walls, signboards, product labels, road signs, markers, etc. Therefore, the inputs have many degrees of variation in font style, orientation, shape, size, color, texture and illumination. The inputs are also subject to camera sensor orientation, location and imperfections causing image blur, pixelation, noise, and geometric and radial distortions. Weather disturbances such as glare, shadow, rain, snow and frost can also greatly affect the performance of STR.

In the body of work on STR, the emphasis has always been on accuracy with little attention paid to speed and computing requirements. In this work, we attempt to put balance on accuracy, speed and efficiency. Accuracy refers to the correctness of recognized text. Speed is measured by how many text images are processed per unit time. Efficiency can be approximated by the number of parameters and computations (eg FLOPS) required to process one image. The number of parameters reflects the memory requirements while FLOPS estimates the number of instructions needed to complete a task. An ideal STR is accurate and fast while requiring only little computing resources.


In the quest to beat the SOTA, most models are zeroing on accuracy with inadequate discussion on the trade off. In order to instill balance on the importance of accuracy, speed and efficiency, we propose to take advantage of the simplicity and efficiency of vision transformers (ViT) \cite{dosovitskiy2020image} such as Data-efficient image Transformer (DeiT) \cite{touvron2020training}. ViT demonstrated that SOTA results in ImageNet \cite{russakovsky2015imagenet} recognition can be achieved using a transformer \cite{vaswani2017attention} encoder only. ViT inherited all the properties of a transformer including its speed and computational efficiency. Using the model weights of DeiT which is simply a ViT trained by knowledge distillation \cite{hinton2015distilling} for better performance, we built an STR that can be trained end-to-end. This  resulted to a simple single stage model architecture that is able to maximize accuracy, speed and computational performance. The tiny version of our ViTSTR achieves 80.3\% accuracy (82.1\% with data augmentation), is fast at 9.3 msec/image, with a small footprint of 5.4M parameters and requires much less computations at 1.3 Giga FLOPS. The small version of ViTSTR achieves a higher accuracy of 82.6\% (84.2\% with data augmentation), is also fast at 9.5 msec/image while requiring 21.5M parameters and 4.6 Giga FLOPS. With data augmentation, the base version of ViTSTR achieves 85.2\% accuracy (83.7\% no augmentation) at 9.8 msec/image but requires 85.8M parameters and 17.6 Giga FLOPS. We adopted the reference \textit{tiny}, \textit{small} and \textit{base} to indicate which ViT/DeiT transformer encoder was used in ViTSTR. As shown in Figure \ref{fig:acc_vs_perf}, almost all versions of our proposed ViTSTR are at or near the frontiers of accuracy vs speed, memory, and computational load indicating optimal trade-offs. To encourage reproducibility, the code of ViTSTR is available at \href{https://github.com/roatienza/deep-text-recognition-benchmark}{https://github.com/roatienza/deep-text-recognition-benchmark}.

\begin{figure}
    \centering

\begin{tabular}{| c | c | c | c | }

\hline
Curved & Uncommon Font Style & Blur and Rotation & Noise\\
 \includegraphics[scale=0.4]{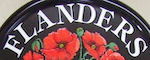} & \includegraphics[scale=0.4]{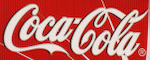} &
 \includegraphics[scale=0.4]{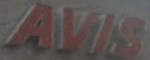} & \includegraphics[scale=0.4]{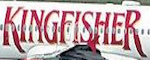}  \\
  \hline
  
   Perspective & Shadow & Occluded and Curved & LowRes \& Pixelation\\
 \includegraphics[scale=0.4]{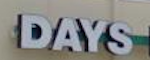} & \includegraphics[scale=0.4]{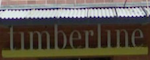} & \includegraphics[scale=0.4]{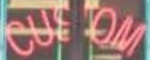} & \includegraphics[scale=0.4]{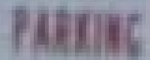}  \\
  \hline
  
\end{tabular}

    \caption{Different variations of text encountered in natural scenes}
    \label{fig:text_variation}
\end{figure}

\tikzstyle{tps}=[draw, fill=blue!20, minimum size=2em]
\tikzstyle{encoder}=[draw, fill=gray!20, minimum size=2em]
\tikzstyle{sequence}=[draw, fill=orange!20, minimum size=2em]
\tikzstyle{prediction}=[draw, fill=green!20, minimum size=2em]
\tikzstyle{decoder}=[draw, fill=yellow!20, minimum size=2em]
\tikzstyle{backbone}=[draw, minimum size=2em]
\tikzstyle{image}=[minimum size=1.5em]

\begin{figure*}[t]
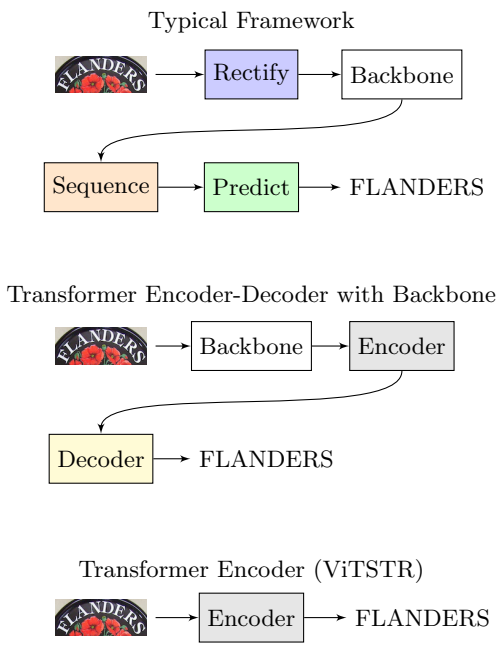

\begin{center}

\begin{tikzpicture}[node distance=2cm,auto,>=latex']
    \node [tps] (tps) {Rectify};
    \node [image] (image) [left of=tps]
    {\includegraphics[width=.1\textwidth]{images/cuve_iitk.png}};
    \node [backbone] (encoder) [right of=tps] {Backbone};
    \node [sequence] (sequence) [below of=image,node distance=1.5cm] {Sequence};
    \node[prediction] (prediction) [right of=sequence] {Predict};
    \node [] (output) [right of=prediction, node distance=2.2cm]{FLANDERS};
    \node [] (label1) [above of=tps, node distance=0.7cm]{Typical Framework};
    \draw[->] (image.east) -- (tps.west);
    \draw[->] (tps.east) -- (encoder.west);
    \draw[->] (encoder.south) .. controls +(down:7mm) and +(up:7mm) ..  (sequence.north);
    \draw[->] (sequence.east) -- (prediction.west);
    \draw[->] (prediction.east) -- (output.west);

    \node [image] (image2)[below of=sequence, node distance=6.5em]     {\includegraphics[width=.1\textwidth]{images/cuve_iitk.png}};
    \node [backbone] (backbone) [right of=image2] {Backbone};
    \node [encoder] (encoder2) [right of=backbone] {Encoder};
    \node[decoder] (decoder) [below of=image2, node distance=1.5cm] {Decoder};
 
    \draw[->] (image2.east) -- (backbone.west);
    \draw[->] (backbone.east) -- (encoder2.west);
    \draw[->] (encoder2.south) .. controls +(down:7mm) and +(up:7mm) ..  (decoder.north);
    \node [] (output2) [right of=decoder, node distance=2.2cm]{FLANDERS};
    \draw[->] (decoder.east) -- (output2.west);
    
    \node [] (label2) [above of=backbone, node distance=0.7cm]{Transformer Encoder-Decoder with Backbone};   

    \node [image] (image3)[below of=decoder, node distance=6.5em]
    {\includegraphics[width=.1\textwidth]{images/cuve_iitk.png}};
    \node [] (coordinate) [right of=image3, node distance=1cm] {};
    \node [encoder] (encoder3) [right of=coordinate, node distance=1cm] {Encoder};
    \node [] (output3) [right of=encoder3, node distance=7em]{FLANDERS};
    \draw[->] (image3.east) -- (encoder3.west);
    \draw[->] (encoder3.east) -- (output3.west);
    \node [] (label3) [above of=encoder3, node distance=1.9em]{Transformer Encoder (ViTSTR)};  

\end{tikzpicture}

\end{center}
   \caption{STR design patterns. Our proposed model, ViTSTR, has the simplest architecture with just one stage.}
\label{fig:str_design_patterns}
\end{figure*}

\section{Related Work}

For machines, reading text in the human environment is a challenging task due to different possible appearances of symbols. Figure \ref{fig:text_variation} shows examples of text in the wild affected by curvature, font style, blur, rotation, noise, geometry, illumination, occlusion and resolution. There are many other factors that could affect text images such as weather condition, camera sensor imperfection, motion, lighting, etc. 

Reading text in natural scenes generally requires two stages: 1) text detection and 2) text recognition. Detection determines the bounding box of the region where text can be found. Once the region is known, text recognition reads the symbols in the image. Ideally, a method is able to do both at the same time. However, the performance of SOTA end-to-end text reading models is still far from modern-day OCR systems and remains an open problem \cite{chen2020text}. In this work, our focus is on text recognition of 96 Latin characters (i.e. 0-9, a-Z, etc.).

STR identifies each character of a text in an image in the correct sequence. Unlike object recognition where usually there is only one category of object, there may be zero or more characters for a given text image. Thus, STR models are more complex. Similar to many vision problems, early methods \cite{neumann2012real,yao2014unified} used hand-crafted features resulting to poor performance. Deep learning has dramatically advanced the field of STR. In 2019, Baek \textit{et al.} \cite{baek2019wrong} presented a framework that models the design patterns of modern STR. Figure \ref{fig:str_design_patterns} shows the four stages or modules of STR. Broadly speaking, even recently proposed methods such as transformer-based models, No-Recurrence sequence-to-sequence Text Recognizer (NRTR) \cite{sheng2019nrtr} and Self-Attention Text Recognition Network (SATRN) \cite{lee2020recognizing} can fit into \textbf{Rectification-Feature Extraction (Backbone)-Sequence Modelling-Prediction} framework.

The Rectification stage removes the distortion from the word image so that the text is horizontal or normalized. This makes it easier for Feature Extraction (Backbone) module to determine invariant features. Thin-Plate-Spline (TPS) \cite{bookstein1989principal} models the distortion by finding and correcting fiducial points. RARE (Robusttext recognizer with Automatic REctification) \cite{shi2016robust}, STAR-Net (SpaTial Attention Residue Network) \cite{liu2016star}, and TRBA (TPS-ResNet-BiLSTM-Attention) \cite{baek2019wrong} use TPS. ESIR (End-to-end trainable Scene text Recognition) \cite{zhan2019esir} employs an iterative rectification network that significantly boosts the performance of text recognition models. In some cases, no rectification is employed such as in CRNN (Convolutional Recurrent Neural Network) \cite{shi2016end}, R2AM (Recursive Recurrent neural networks with Attention Modeling) \cite{lee2016recursive}, GCRNN (Gated Recurrent Convolution Neural Network) \cite{wang2017gated} and Rosetta \cite{borisyuk2018rosetta}.

The role of Feature Extraction (Backbone) stage is to automatically determine the invariant features of each character symbol. STR uses the same feature extractors in object recognition tasks such as VGG \cite{simonyan2014very}, ResNet \cite{he2016deep}, and a variant of CNN called RCNN \cite{lee2016recursive}. Rosetta,  STAR-Net and TRBA use ResNet. RARE and CRNN  extract features using VGG. R2AM and GCRNN build on RCNN. Transformer-based models NRTR and SATRN use customized CNN blocks to extract features for transformer encoder-decoder text recognition.

Since STR is a multi-class sequence prediction, there is a need to remember long-term dependency. The role of Sequence modelling such as BiLSTM is to make a consistent context between the current character features and the past/future characters features. CRNN, GRCNN, RARE, STAR-Net and TRBA use BiLSTM. Other models such as Rosetta and R2AM do not employ sequence modelling to speed up prediction.

The Prediction stage examines the features resulting from the Backbone or Sequence modelling to arrive at a sequence of characters prediction. CTC (Connectionist Temporal Classification) \cite{graves2006connectionist}  maximizes the likelihood of an output sequence by efficiently summing over all possible input-output sequence alignments \cite{chen2020text}. Alternative to CTC is Attention Mechanism \cite{bahdanau2014neural} that learns the alignment between the image features and symbols. CRNN, GRCNN, Rosetta and STAR-Net use CTC. R2AM, RARE and TRBA are Attention-based.

Like in natural language processing (NLP), transformers overcome sequence modelling and prediction by doing parallel self-attention and prediction. This resulted to a fast and efficient model. As shown in Figure \ref{fig:str_design_patterns}, current transformer-based STR models still require a Backbone and a Transformer Encoder-Decoder. Recently, ViT \cite{dosovitskiy2020image} proved that it is possible to beat the performance of deep networks such as ResNet \cite{he2016deep} and EfficientNet \cite{tan2019efficientnet} on ImageNet1k \cite{russakovsky2015imagenet} classification by using the transformer encoder only but pre-training it on very large datasets such as ImageNet21k and JFT-300M. DeiT \cite{touvron2020training} demonstrated that ViT does not need a large dataset and can even achieve better results but it must be trained using knowledge distillation \cite{hinton2015distilling}. ViT, using pre-trained weights of DeiT, is the basis of our proposed fast and efficient STR called ViTSTR. As shown in Figure \ref{fig:str_design_patterns}, ViTSTR is a very simple model with just one stage that can easily halve the number of parameters and FLOPS of a transformer-based STR.

\tikzstyle{encoder}=[draw, fill=gray!20, minimum width=26em, minimum height=3em]
\tikzstyle{linear}=[draw, fill=teal!20, minimum width=23em, minimum height=3em]
\tikzstyle{posit}=[draw, fill=yellow!20, minimum width=1.05em, minimum height=1.35em]
\tikzstyle{box}=[draw, fill=teal!20, minimum width=1.05em, minimum height=1.35em]
\tikzstyle{image}=[minimum size=1.6em]
\tikzstyle{thumb}=[minimum size=.4em]

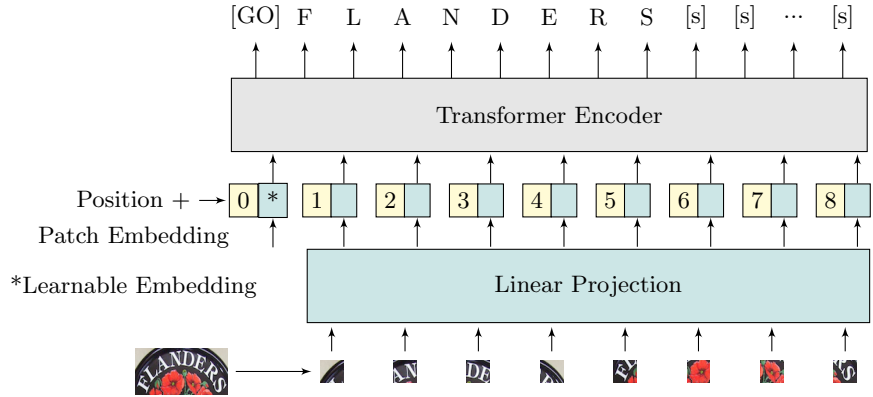
\begin{figure*}[t]
\begin{center}

\begin{tikzpicture}[node distance=2cm,auto,>=latex']
    \node [encoder] (encoder) [] {Transformer Encoder};
    
    \node [linear] (linear) [below of=encoder, xshift=1.6em, node distance=7em] {Linear Projection};

    \node [posit] (pos1) [below of=encoder, node distance=3.5em, xshift=-12.5em] {0};
    \node [box] (box1) [below of=encoder, node distance=3.5em, xshift=-11.3em] {*};
    
    \node [posit] (pos2) [below of=encoder, node distance=3.5em, xshift=-9.5em] {1};
    \node [box] (box2) [below of=encoder, node distance=3.5em, xshift=-8.4em] {};
    
    \node [posit] (pos3) [below of=encoder, node distance=3.5em, xshift=-6.5em] {2};
    \node [box] (box3) [below of=encoder, node distance=3.5em, xshift=-5.4em] {};
    
    \node [posit] (pos4) [below of=encoder, node distance=3.5em, xshift=-3.5em] {3};
    \node [box] (box4) [below of=encoder, node distance=3.5em, xshift=-2.4em] {};
    
    \node [posit] (pos5) [below of=encoder, node distance=3.5em, xshift=-.5em] {4};
    \node [box] (box5) [below of=encoder, node distance=3.5em, xshift=.6em] {};    
    
    \node [posit] (pos6) [below of=encoder, node distance=3.5em, xshift=2.5em] {5};
    \node [box] (box6) [below of=encoder, node distance=3.5em, xshift=3.6em] {};  
    
    \node [posit] (pos7) [below of=encoder, node distance=3.5em, xshift=5.5em] {6};
    \node [box] (box7) [below of=encoder, node distance=3.5em, xshift=6.6em] {};  
    
    \node [posit] (pos8) [below of=encoder, node distance=3.5em, xshift=8.5em] {7};
    \node [box] (box8) [below of=encoder, node distance=3.5em, xshift=9.6em] {};  
    
    \node [posit] (pos9) [below of=encoder, node distance=3.5em, xshift=11.5em] {8};
    \node [box] (box9) [below of=encoder, node distance=3.5em, xshift=12.6em] {};  
    
    \node [thumb] (thumb1) [below of=linear, node distance=3.5em, xshift=-10.5em]
    {\includegraphics[width=.025\textwidth]{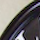}};
    
    \node [image] (image) [left of=thumb1, node distance=2cm]
    {\includegraphics[width=.1\textwidth]{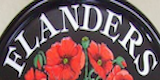}};
    
    \node [thumb] (thumb2) [below of=linear, node distance=3.5em, xshift=-7.5em]
    {\includegraphics[width=.025\textwidth]{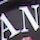}};
    
    \node [thumb] (thumb3) [below of=linear, node distance=3.5em, xshift=-4.5em]
    {\includegraphics[width=.025\textwidth]{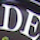}};
    
    \node [thumb] (thumb4) [below of=linear, node distance=3.5em, xshift=-1.5em]
    {\includegraphics[width=.025\textwidth]{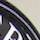}};

    \node [thumb] (thumb5) [below of=linear, node distance=3.5em, xshift=1.5em]
    {\includegraphics[width=.025\textwidth]{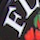}};
    
    \node [thumb] (thumb6) [below of=linear, node distance=3.5em, xshift=4.5em]
    {\includegraphics[width=.025\textwidth]{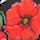}};    
    
    \node [thumb] (thumb7) [below of=linear, node distance=3.5em, xshift=7.5em]
    {\includegraphics[width=.025\textwidth]{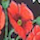}};
    
    \node [thumb] (thumb8) [below of=linear, node distance=3.5em, xshift=10.5em]
    {\includegraphics[width=.025\textwidth]{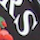}};
    
    \draw[->] (-12em,1.5em) -- +(up:1.5em); 
    \node[] at (-12em,4em) {[GO]};

    \draw[->] (-10em,1.5em) -- +(up:1.5em); 
    \node[] at (-10em,4em) {F};
    
    \draw[->] (-8em,1.5em) -- +(up:1.5em); 
    \node[] at (-8em,4em) {L};
    
    \draw[->] (-6em,1.5em) -- +(up:1.5em); 
    \node[] at (-6em,4em) {A};
    
    \draw[->] (-4em,1.5em) -- +(up:1.5em); 
    \node[] at (-4em,4em) {N};
    
    \draw[->] (-2em,1.5em) -- +(up:1.5em); 
    \node[] at (-2em,4em) {D};
    
    \draw[->] (0em,1.5em) -- +(up:1.5em); 
    \node[] at (0em,4em) {E};
    
    \draw[->] (2em,1.5em) -- +(up:1.5em); 
    \node[] at (2em,4em) {R};
    
    \draw[->] (4em,1.5em) -- +(up:1.5em); 
    \node[] at (4em,4em) {S};
    
    \draw[->] (6em,1.5em) -- +(up:1.5em); 
    \node[] at (6em,4em) {[s]};

    \draw[->] (8em,1.5em) -- +(up:1.5em); 
    \node[] at (8em,4em) {[s]};

    \draw[->] (10em,1.5em) -- +(up:1.5em); 
    \node[] at (10em,4em) {...};
    
    \draw[->] (12em,1.5em) -- +(up:1.5em); 
    \node[] at (12em,4em) {[s]};
    
    \draw[->] (image.east) -- (thumb1.west);
    \draw[->] (thumb1.north) -- +(up:1em);  
    \draw[->] (thumb2.north) -- +(up:1em);
    \draw[->] (thumb3.north) -- +(up:1em);
    \draw[->] (thumb4.north) -- +(up:1em);
    \draw[->] (thumb5.north) -- +(up:1em);
    \draw[->] (thumb6.north) -- +(up:1em);
    \draw[->] (thumb7.north) -- +(up:1em);
    \draw[->] (thumb8.north) -- +(up:1em);
    
    \draw[->] (box1.north) -- +(up:1.2em); 
    \draw[<-] (box1.south) -- +(down:1.2em);
    
    \draw[->] (box2.north) -- +(up:1.2em); 
    \draw[<-] (box2.south) -- +(down:1.2em);
    
    \draw[->] (box3.north) -- +(up:1.2em); 
    \draw[<-] (box3.south) -- +(down:1.2em);
    
    \draw[->] (box4.north) -- +(up:1.2em); 
    \draw[<-] (box4.south) -- +(down:1.2em);
    
    \draw[->] (box5.north) -- +(up:1.2em); 
    \draw[<-] (box5.south) -- +(down:1.2em);

    \draw[->] (box6.north) -- +(up:1.2em); 
    \draw[<-] (box6.south) -- +(down:1.2em);
    
    \draw[->] (box7.north) -- +(up:1.2em); 
    \draw[<-] (box7.south) -- +(down:1.2em);
    
    \draw[->] (box8.north) -- +(up:1.2em); 
    \draw[<-] (box8.south) -- +(down:1.2em);
    
    \draw[->] (box9.north) -- +(up:1.2em); 
    \draw[<-] (box9.south) -- +(down:1.2em);
    
    \node[] (patch1)[left of=pos1, node distance=4.5em]{Position +};
    \node[] (patch2)[below of=patch1, node distance=1.5em]{Patch Embedding};
    
    \node[] (patch3)[below of=patch2, node distance=2em]{*Learnable Embedding};
    
    \draw[->] (patch1.east) -- +(right:1.1em); 
    
    
\end{tikzpicture}

\end{center}
   \caption{Network architecture of ViTSTR. An input image is first converted into patches. The patches are converted into 1D vector embeddings (flattened 2D patches). As input to the encoder, a learnable patch embedding is added together with a position encoding for each embedding. The network is trained end-to-end to predict a sequence of characters. [GO] is a pre-defined start of sequence symbol while [s] represents a space or end of a character sequence.}
\label{fig:vitstr}
\end{figure*}

\tikzstyle{encoder}=[draw, fill=gray!20, minimum width=13em, minimum height=24em]
\tikzstyle{patch}=[draw, fill=teal!20, minimum width=10em, minimum height=2.5em]
\tikzstyle{norm}=[draw, fill=orange!20, minimum width=10em, minimum height=2.5em]
\tikzstyle{attention}=[draw, fill=green!20, minimum width=10em, minimum height=2.5em, text width=9em,align=center ]
\tikzstyle{plus}=[draw, circle, minimum size=1em, fill=white]
\tikzstyle{mlp}=[draw, fill=red!20, minimum width=10em, minimum height=2.5em]
\tikzstyle{box}=[draw, minimum width=1.05em, minimum height=1.35em]

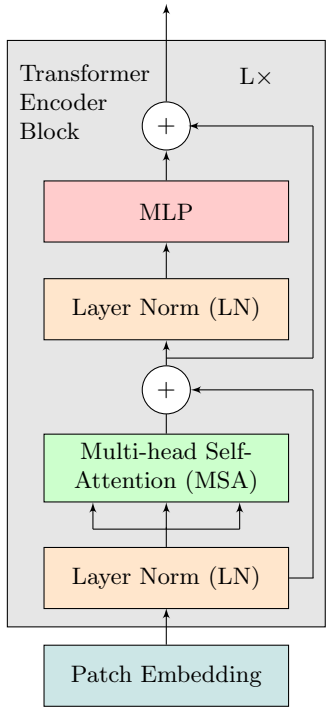
\begin{figure*}[t]
\begin{center}

\begin{tikzpicture}[node distance=2cm,auto,>=latex']
    \node[encoder] (encoder) [] {};
    \node[patch] (patch)[below of=encoder, node distance=14em] {Patch Embedding};
    \draw[->] (patch.north) -- +(up: 1.5em);
    \node[norm] (norm)[below of=encoder, node distance=10em] {Layer Norm (LN)};
    \node[attention] (attention)[below of=encoder, node distance=5.5em] {Multi-head Self-Attention (MSA)};
    \draw[-] (norm.north) -- +(up: 0.8em);
    \draw[-] (-3em,-8em) -- +(6em, 0em);
    \draw[->] (-3em,-8em) -- +(0em, 1.1em);
    \draw[->] (0em,-8em) -- +(0em, 1.1em);
    \draw[->] (3em,-8em) -- +(0em, 1.1em);
    \node[plus] (plus)[below of=encoder, node distance=2.3em] {+};
    \draw[-] (attention.north) -- +(0em, 0.8em);
    \draw[-] (norm.east) -- +(1em,0em);
    \draw[-] (6em,-10em) -- +(0em,7.7em);
    \draw[->] (6em,-2.3em) -- +(-5em,0em);
    \node[norm] (norm1)[above of=encoder, node distance=1em] {Layer Norm (LN)};
    \draw[->] (plus.north) -- (norm1.south);
    \node[mlp] (mlp)[above of=encoder, node distance=5em] {MLP};
    \node[plus] (plus1)[above of=encoder, node distance=8.5em] {+};
    \node[] (label1)[above of=plus1, xshift=-3.5em, node distance=1em, text width=5em]{Transformer Encoder Block};
    \node[] (label2)[above of=plus1, xshift=5.5em, node distance=2em, text width=5em]{L$\times$};
    \draw[->] (norm1.north)--(mlp.south);
    \draw[->] (mlp.north) -- (plus1.south);
    \draw[->] (plus1.north) -- +(0em, 4em);
    \draw[-] (0em, -1em) -- +(6em,0em);
    \draw[-] (6em, -1em) -- +(0em, 9.5em);
    \draw[->] (6em, 8.5em) -- +(-5.1em,0em);
    
\end{tikzpicture}

\end{center}
   \caption{A transformer encoder is a stack of $L$ identical encoder blocks.}
\label{fig:encoder}
\end{figure*}    

\section{Vision Transformer for STR}

Figure \ref{fig:vitstr} shows the model architecture of ViTSTR in detail. The only difference between ViT and ViTSTR is the prediction head. Instead of single object-class recognition, ViTSTR must identify multiple characters with the correct sequence order and length. The prediction is done in parallel.

The ViT model architecture is similar to the original transformer by Vaswani \textit{et al.} \cite{vaswani2017attention}. The difference is only the encoder part is utilized. The original transformer was designed for NLP tasks. Instead of word embeddings, each input image $\textbf{x}\in{	\mathbb{R}}^{H\times{W}\times{C}}$ is reshaped into a sequence of flattened 2D patches $\textbf{x}^{p}\in{	\mathbb{R}}^{N\times{P^2}{C}}$. The image dimension is $H\times{W}$ with $C$ channels while the patch dimension is $P\times{P}$. The resulting patch sequence length is $N$. The transformer encoder uses a constant width $D$ for embedding and features in all its layers. To match this size, each flattened patch is converted to an embedding of size $D$ via linear projection. This is shown as small boxes with teal color in Figure \ref{fig:vitstr}. 

A learnable class embedding of the same dimension $D$ is prepended with the sequence. A unique position encoding of the same dimension $D$ is added to each embedding. The resulting vector sum is the input to the encoder. In ViTSTR, a learnable position encoding is used.

In the original ViT, the output vector corresponding to the learnable class embedding is used for object category prediction. In ViTSTR, this corresponds to the [GO] token. Furthermore, instead of just extracting one output vector, we extract multiple feature vectors from the encoder. The number is equal to the maximum length of text in our dataset plus two for the [GO] and [s] tokens. We use the [GO] token to mark the beginning of the text prediction and [s] to indicate the end or a space. [s] is repeated at the end of each text prediction up to the maximum sequence length to mark that nothing follows after the text characters.

Figure \ref{fig:encoder} shows the layers inside one encoder block. Every input goes through Layer Normalization (LN). The Multi-head Self-Attention layer (MSA) determines the relationships between feature vectors. Vaswani \textit{et al.} \cite{vaswani2017attention} found out that using multiple heads instead of just one allows the model to jointly attend to information from different representation subspaces at different positions. The number of heads is $H$. The Multilayer Perceptron (MLP) performs feature extraction. Its input is also layer normalized. The MLP is made of 2 layers with GELU activation \cite{hendrycks2016gaussian}. Residual connection is placed between the output of LN and MSA/MLP.

In summary, the input to the encoder is:

\begin{equation}
\label{eq:encoder_input}
\textbf{z}_{0}=[\textbf{x}_{class};\textbf{x}^{1}_{p}\textbf{E};\textbf{x}^{2}_{p}\textbf{E};...;\textbf{x}^{N}_{p}\textbf{E}] + \textbf{E}_{pos},    
\end{equation}

where $\textbf{E}\in\mathbb{R}^{P^2C\times{D}}$ and $\textbf{E}_{pos}\in\mathbb{R}^{ (N+1)\times{D} }$.

The output of MSA block is:

\begin{equation}
\label{eq:msa}
\textbf{z}^{'}_{l}=MSA(LN(\textbf{z}_{l-1})) + \textbf{z}_{l-1},     
\end{equation}

for $l=1...L$. $L$ is the depth or the number of encoder blocks. A transformer encoder is made of a stack of $L$ encoder blocks. 

The output of the MLP block is:

\begin{equation}
\label{eq:mlp}
\textbf{z}_{l}=MLP(LN(\textbf{z}^{'}_{l})) + \textbf{z}^{'}_{l},  
\end{equation}

for $l=1...L$.

Finally, the head is made of a sequence of linear projections forming the word prediction:

\begin{equation}
   \label{eq:head}
\textbf{y}_{i}=Linear(\textbf{z}^{i}_{L}), 
\end{equation}

for $i=1...S$. $S$ is the maximum text length plus two for [GO] and [s] tokens. Table \ref{tab:vitstr_config} summarizes the ViTSTR configurations.

\begin{table}
\caption{ViTSTR configurations}
\label{tab:vitstr_config}

\begin{center}
\begin{tabular}{|l | c | c | c | c | c|}

\hline

ViTSTR  & Patch Size & Depth & Embedding Size  & No. of Heads &  Seq Length\\
Version  & $P$ & $L$ & $D$  & $H$ &  $S$\\
\hline
Tiny & 16 & 12 & 192 & 3 & 27  \\
\hline
Small & 16 & 12 & 384 & 6 & 27  \\
\hline
Base & 16 & 12 & 768 & 12 & 27  \\
\hline

\end{tabular}
\end{center}
\end{table}

\section{Experimental Results and Discussion}

In order to evaluate different strong baseline STR methods, we used the framework developed by Baek \textit{et al.} \cite{baek2019wrong}. A unified framework is important in order to arrive at a fair evaluation of different models. A unified framework ensures consistent train and test conditions are used in the evaluation. Following discussion describes the train and test datasets which have been the point of contention in performance comparisons. Using different train and test datasets can heavily tilt in favor or against a certain performance reporting.

After discussing the train and test datasets, we present the evaluation and analysis across different models using the unified framework. 

\subsection{Train Dataset}

\begin{figure}
    \centering
\begin{tabular}{| c | c  |}
\hline
MJ & ST \\
\hline
\includegraphics[scale=0.6]{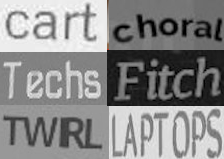} & \includegraphics[scale=0.6]{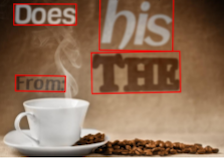} \\
\hline

\end{tabular}    
    \caption{Samples from datasets with synthetic images.}
    \label{fig:synth_dataset}
\end{figure}

Due to the lack of a big dataset of real data, the practice in STR model training is to use synthetic data. Two popular datasets are used: 1) MJSynth (MJ) \cite{jaderberg2014synthetic} or also known as Synth90k and 2) SynthText (ST) \cite{gupta2016synthetic}. 

\textbf{MJSynth (MJ)} is a synthetically generated dataset made of 8.9M realistically looking  words images. MJSynth was designed to have 3 layers: 1) background, 2) foreground and 3) optional shadow/border. It uses 1,400 different fonts. The font kerning, weight, underline and other properties are varied. MJSynth also utilizes different background effects, border/shadow rendering, base coloring, projective distortion, natural image blending and noise.

\textbf{SynthText (ST)} is another synthetically generated dataset made of 5.5M word images. SynthText was generated by blending synthetic text on natural images. It uses the scene geometry, texture, and surface normal to naturally blend and distort a text rendering on the surface of an object within the image. Similar to MJSynth, SynthText uses random fonts for its text. The word images were cropped from the natural images embedded with synthetic text.

In the STR framework, each dataset contributes 50\% to the total train dataset. Combining 100\% of both datasets resulted to performance deterioration \cite{baek2019wrong}. Figure \ref{fig:synth_dataset} shows sample images from MJ and ST.

\subsection{Test Dataset}

\begin{figure}
    \centering
    
\begin{center}

\begin{tabular}{| r | c  c  c  |  r | c  c  c |}

\hline
\multicolumn{4}{|c|}{\textbf{Regular Dataset} } & \multicolumn{4}{|c|}{\textbf{Irregular Dataset} } \\
\hline
IIIT5K 
&
\includegraphics[scale=0.28]{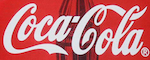} & 
\includegraphics[scale=0.28]{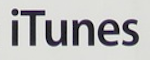} &
\includegraphics[scale=0.28]{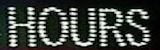}
&
IC15
&

\includegraphics[scale=0.28]{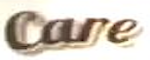}
&
\includegraphics[scale=0.28]{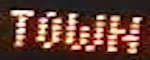}
&
\includegraphics[scale=0.28]{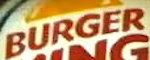}

\\ 

SVT 
&
\includegraphics[scale=0.28]{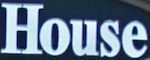} 
& 
\includegraphics[scale=0.28]{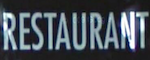} 
&
\includegraphics[scale=0.28]{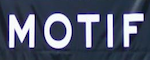}

&
SVTP 
&
\includegraphics[scale=0.28]{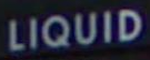} & 
\includegraphics[scale=0.28]{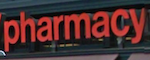} &
\includegraphics[scale=0.28]{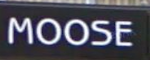}

\\

IC03 
&
\includegraphics[scale=0.28]{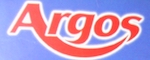} & 
\includegraphics[scale=0.28]{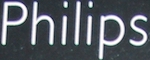} &
\includegraphics[scale=0.28]{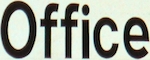}
&
CT 
&
\includegraphics[scale=0.28]{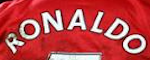} 
& \includegraphics[scale=0.28]{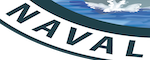} 
&
\includegraphics[scale=0.28]{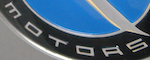}

\\ 

IC13 
&
\includegraphics[scale=0.28]{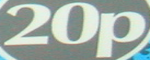} & 
\includegraphics[scale=0.28]{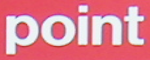} &
\includegraphics[scale=0.28]{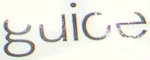}
& 
{}
& 
{}
&
{}
&
{}
\\ 

\hline

\end{tabular}
\end{center}
    \caption{Samples from datasets with real images.}
    \label{fig:real_dataset}
\end{figure}

The test dataset is made of several small publicly available STR datasets of text in natural images. These datasets are generally group into two: 1) Regular and 2) Irregular. 

The regular datasets have text images that are frontal, horizontal and have minimal amount of distortion. IIIT5K-Words \cite{mishra2012scene}, Street View Text (SVT) \cite{wang2011end}, ICDAR2003 (IC03) \cite{lucas2005icdar} and ICDAR2013 (IC13) \cite{karatzas2013icdar} are considered regular datasets. Meanwhile, irregular datasets contain text with challenging appearances such curved, vertical, perspective, low-resolution or distorted. ICDAR2015 (IC15) \cite{karatzas2015icdar}, SVT Perspective (SVTP) \cite{phan2013recognizing} and CUTE80 (CT) \cite{risnumawan2014robust} belong to irregular datasets. Figure \ref{fig:real_dataset} shows samples from regular and irregular datasets. For both datasets, only the test splits are used for the evaluation.

\begin{table}[]
    \centering
    \caption{Train conditions}
    \begin{tabular}{|l|l|}
    \hline
        \textbf{Train dataset}: 50\%MJ + 50\%ST & \textbf{Batch size}: 192\\
    \hline
        \textbf{Epochs}: 300 & \textbf{Parameter initialization}:  He \cite{he2015delving} \\
    \hline
         \textbf{Optimizer}: Adadelta \cite{zeiler2012adadelta} & \textbf{Learning rate}: 1.0\\
    \hline
         \textbf{Adadelta $\rho$}: 0.95 & \textbf{Adadelta $\epsilon$}: $1e^{-8}$\\
    \hline
        \textbf{Loss}: Cross-Entropy/CTC & \textbf{Gradient clipping}: 5.0 \\
    \hline
        \textbf{Image size}: $100\times{32}$ & \textbf{Channels}: 1 (grayscale) \\
    \hline
    \end{tabular}
    \label{tab:train_condition}
\end{table}

\subsubsection{Regular Dataset}
\begin{itemize}
  \item \textbf{IIIT5K} contains 3,000 images for testing. The images are mostly from street scenes such as sign board, brand logo, house number or street sign. 

  \item \textbf{SVT} has 647 images for testing. The text images are cropped from Google Street View images.
  
  \item \textbf{IC03} contains 1,110 test images from ICDAR2003 Robust Reading Competition. Images were captured from natural scenes. After removing words that are less than 3 characters in length, the result is 860 images. However, 7 additional images were found to be missing. Hence, the framework also contains the 867 test images version. 
  
  \item \textbf{IC13} is an extension of IC03 and shares similar images. IC13 was created for the ICDAR2013 Robust Reading Competition.  In the literature and in the framework, two versions of the test dataset are used: 1) 857 and 2) 1,015.
\end{itemize}

\subsubsection{Irregular Dataset}

\begin{itemize}
    \item \textbf{IC15} has text images for the ICDAR2015 Robust Reading Competition. Many images are blurry, noisy, rotated, and sometimes of low-resolution since these were captured using Google Glasses with the wearer undergoing unconstrained motion. Two versions are used in the literature and in the framework: 1) 1,811 and 2) 2,077 images. The 2,077 version contains rotated, vertical, perspective-shifted and curved images. 
    \item \textbf{SVTP} has 645 test images from Google Street View. Most are images of business signage.
    
    \item \textbf{CT} focuses on curved text images captured from shirts and product logos. The dataset has 288 images. 
\end{itemize}

\subsection{Experimental Setup}

The recommended training configurations in the framework are listed in Table \ref{tab:train_condition}. We reproduced the results of several strong baseline models: CRNN, R2AM, GCRNN, Rosetta, RARE, STAR-Net and TRBA for a fair comparison with ViTSTR. We trained all models for at least 5 times using different random seeds. The best performing weights on the test datasets are saved to get the mean evaluation scores. 

For ViTSTR, we used the same train configurations except that the input is resized to $224\times{224}$ to match the dimension of the pre-trained DeiT \cite{touvron2020training}. The pre-trained weights file of DeiT is automatically downloaded before training ViTSTR. ViTSTR can be trained end-to-end with no parameters frozen.

Tables \ref{tab:model_acc} and \ref{tab:model_speed} show the performance scores of different models. We report the accuracy, speed, number of parameters and FLOPS to get the overall picture of trade-offs as shown in Figure \ref{fig:acc_vs_perf}. For accuracy, we follow the framework evaluation protocol in most STR models of case sensitive training and case insensitive evaluation. For speed, the reported numbers are based on model run time on a 2080Ti GPU. Unlike in other model benchmarks such as in \cite{li2019show,litman2020scatter}, we do not rotate vertical text images (e.g. Table \ref{tab:failed_cases} IC15) before evaluation.

\begin{table}[]
    \centering
    \caption{Model accuracy. Bold: highest for all, Underscore: highest no augmentation.}
    \begin{tabular}{l |c  c  c  c  c  c  c  c  c  c  c  c }
    \hline
    Model & IIIT & SVT & \multicolumn{2}{c}{IC03} & \multicolumn{2}{c}{IC13} & \multicolumn{2}{c}{IC15} & SVTP & CT & Acc & Std \\
    
    {} & 3000 & 647 & 860 & 867 & 857 & 1015 & 1811 & 2077 & 645 & 288 & \% & {} \\
    \hline
    
    CRNN \cite{shi2016end} & 81.8	&80.1	&91.7	&91.5	&89.4	&88.4	&65.3	&60.4	&65.9	&61.5	&76.7	&0.3 \\
    
    R2AM \cite{lee2016recursive}& 83.1	&80.9	&91.6	&91.2	&90.1	&88.1	&68.5	&63.3	&70.4	&64.6	&78.4	&0.9 \\
    
    GCRNN \cite{wang2017gated}& 82.9	&81.1	&92.7	&92.3	&90.0	&88.4	&68.1	&62.9	&68.5	&65.5	&78.3	&0.1 \\
    
    Rosetta \cite{borisyuk2018rosetta}& 82.5	&82.8	&92.6	&91.8	&90.3	&88.7	&68.1	&62.9	&70.3	&65.5	&78.4	&0.4 \\
    
    RARE \cite{shi2016robust}& 86.0	&85.4	&93.5	&93.4	&92.3	&91.0	&73.9	&68.3	&75.4	&71.0	&82.1	&0.3\\
    
    STAR-Net \cite{liu2016star}& 	85.2	&84.7	&93.4	&93.0	&91.2	&90.5	&74.5	&68.7	&74.7	&69.2	&81.8	&0.1 \\
    
    TRBA \cite{baek2019wrong}& \underline{87.8}	& \underline{87.6}	& \underline{94.5}	& \underline{94.2}	& \textbf{93.4}	& \underline{92.1}	& \underline{77.4}	& \underline{71.7}	& 78.1	& \underline{75.2}	& \underline{84.3}	& 0.1\\
    
    ViTSTR-Tiny	&83.7	&83.2	&92.8	&92.5	&90.8	&89.3	&72.0	&66.4	&74.5	&65.0	&80.3	&0.2\\
    
    ViTSTR-Tiny+Aug	&85.1	&85.0	&93.4	&93.2	&90.9	&89.7	&74.7	&68.9	&78.3	&74.2	&82.1	&0.1\\
   
    ViTSTR-Small & 85.6	&85.3	&93.9	&93.6	&91.7	&90.6	&75.3	&69.5	&78.1	&71.3	&82.6	&0.3 \\
    
    ViTSTR-Small+Aug & 86.6	&87.3	&94.2	&94.2	&92.1	&91.2	&77.9	&71.7	&81.4	&77.9	&84.2	&0.1\\
    
    ViTSTR-Base &86.9	&87.2	&93.8	&93.4	&92.1	&91.3	&76.8	&71.1	&\underline{80.0}	&74.7	&83.7	&0.1\\
    
    ViTSTR-Base+Aug &\textbf{88.4}	&\textbf{87.7}	&\textbf{94.7}	&\textbf{94.3}	&93.2	&\textbf{92.4}	&\textbf{78.5}	&\textbf{72.6}	&\textbf{81.8}	&\textbf{81.3}	&\textbf{85.2}	&0.1\\
    
    \hline
    \end{tabular}

    \label{tab:model_acc}
\end{table}

\begin{table}[]
    \centering
    \caption{Model accuracy, speed, and computational requirements on a 2080Ti GPU.}
    \begin{tabular}{l|c c c c}
    \hline
     Model & Accuracy & Speed & Parameters & FLOPS  \\
     {} & \% & msec/image & $1\times{10^6}$ & $1\times{10^9}$ \\
     \hline
    CRNN \cite{shi2016end}& 76.7 & 3.7 & 8.5 & 1.4 \\
    R2AM \cite{lee2016recursive}& 78.4 & 22.9 & 2.9 & 2.0 \\
    GRCNN \cite{wang2017gated}& 78.3 & 11.2 & 4.8 & 1.8 \\
    Rosetta \cite{borisyuk2018rosetta}& 78.4 & 5.3 & 44.3 & 10.1\\
    RARE \cite{shi2016robust}& 82.1 & 18.8 & 10.8 & 2.0\\
    STAR-Net \cite{liu2016star}& 81.8 & 8.8 & 48.9 & 10.7\\
    TRBA \cite{baek2019wrong}& 84.3 & 22.8 & 49.6 & 10.9\\
    ViTSTR-Tiny & 80.3 & 9.3 & 5.4 & 1.3\\
    ViTSTR-Tiny+Aug & 82.1 & 9.3 & 5.4 & 1.3\\
    ViTSTR-Small & 82.6 & 9.5 & 21.5 & 4.6\\
    ViTSTR-Small+Aug & 84.2 & 9.5 & 21.5 & 4.6\\
    
    ViTSTR-Base & 83.7 & 9.8 & 85.8 & 17.6\\
    ViTSTR-Base+Aug & 85.2 & 9.8 & 85.8 & 17.6\\
    \hline     
    \end{tabular}

    \label{tab:model_speed}
\end{table}

\begin{figure}
    \centering
    
    \begin{tabular}{| c | c |  c | c | c |}

    \hline
    Original & Invert & Curve & Blur & Noise\\
    \includegraphics[scale=0.65]{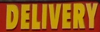} & \includegraphics[scale=0.65]{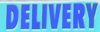} &
    \includegraphics[scale=0.65]{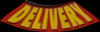} &
    \includegraphics[scale=0.65]{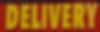} &
    \includegraphics[scale=0.65]{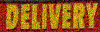}\\ 
    \hline
    Distort & Rotate & Stretch/Comp. & Perspective & Shrink\\
    \includegraphics[scale=0.65]{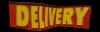} & \includegraphics[scale=0.65]{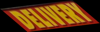} &
    \includegraphics[scale=0.65]{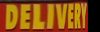} &
    \includegraphics[scale=0.65]{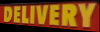} &
    \includegraphics[scale=0.65]{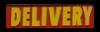}\\ 
    \hline
    \end{tabular}
    \caption{Illustration of data augmented text images designed for STR.}
    \label{fig:data_augmentation}
\end{figure}

\begin{figure}
    \centering
    
    \begin{tabular}{| c | c |  c | c | c | c |}

    \hline
    N & e & s & t & l & e\\
    \includegraphics[scale=0.27]{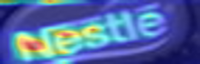} & \includegraphics[scale=0.27]{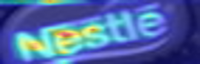} &
    \includegraphics[scale=0.27]{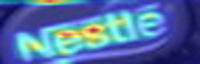} &
    \includegraphics[scale=0.27]{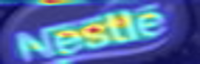} &
    \includegraphics[scale=0.27]{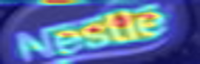} & 
    \includegraphics[scale=0.27]{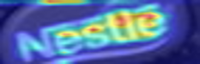}\\ 
    \hline
    \end{tabular}
    \caption{ViTSTR attention as it reads out \textbf{Nestle} text image.}
    \label{fig:attention}
\end{figure}

\subsection{Data Augmentation}
Using a recipe of data augmentation specifically targeted for STR can significantly boost the accuracy of ViTSTR. In Figure \ref{fig:data_augmentation}, we can see how different data augmentations alter the image but not the meaning of text within. Table \ref{tab:model_acc} shows that applying RandAugment \cite{cubuk2020randaugment} on different image transformations such as inversion, curving, blur, noise, distortion, rotation, stretching/compressing, perspective, and shrinking  improved the generalization of ViTSTR-Tiny by +1.8\%, ViTSTR-Small by +1.6\% and ViTSTR-Base by 1.5\%. The biggest increase in accuracy is on irregular datasets such as CT (+9.2\% tiny, +6.6\% small and base), SVTP (+3.8\% tiny, +3.3\% small, +1.8\% base), IC15 1,811 (+2.7\% tiny, +2.6\% small, +1.7\% base) and IC15 2,077 (+2.5\% tiny, +2.2\% small, +1.5\% base).

\begin{table}[]
    \centering
    \caption{ViTSTR sample failed prediction from each test dataset. From first to last row: input image, ground truth, prediction, dataset. Wrong symbol prediction in \textcolor{red}{red}.}
    \begin{tabular}{ c c  c  c  c  c  c }
 
    \includegraphics[scale=0.43]{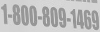} &
    \includegraphics[scale=0.43]{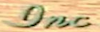} &
    \includegraphics[scale=0.43]{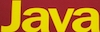} &
    \includegraphics[scale=0.43]{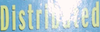} &
    \includegraphics[scale=0.43]{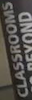} &
    \includegraphics[scale=0.43]{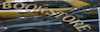} &
    \includegraphics[scale=0.43]{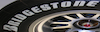} \\
    \footnotesize{18008091469} & INC & JAVA & \scriptsize{Distributed} & \tiny{CLASSROOMS} &  \scriptsize{BOOKSTORE} & \tiny{BRIDGESTONE} \\ 
    \scriptsize{1800\textcolor{red}{B}09\textcolor{red}{4}46\textcolor{red}{Y}} & \textcolor{red}{O}nc & \textcolor{red}{I}AVA & \scriptsize{Distrib\textcolor{red}{a}ted} & \textcolor{red}{\scriptsize{Io-14DD07}} & \scriptsize{BOOKSTOR\textcolor{red}{A}} & \scriptsize{\textcolor{red}{Dueeesrreee}} \\
    
    \hline
    IIIT5K & SVT & IC03 & IC13 & IC15 & SVTP & CUTE80 \\ 
    \end{tabular}

    \label{tab:failed_cases}
\end{table}

\subsection{Attention}
Figure \ref{fig:attention} shows the attention map of ViTSTR as it reads out a text image. While the attention is properly focused on each character, ViTSTR also pays attention to neighboring characters. Perhaps, a context is placed during individual symbol prediction.

\subsection{Performance Penalty}
Every time a stage in an STR model is added, there is a gain in accuracy but at a cost of slower speed and bigger computational requirements. For example, RARE$\hookrightarrow$TRBA increases the accuracy by 2.2\% but requires 38.8M more parameters and slows down the task completion by 4 msec/image. Replacing the CTC stage by Attention like in STAR-Net$\hookrightarrow$TRBA significantly slows down the computation from 8.8 msec/image to 22.8 msec/image to gain an additional 2.5\% in accuracy. In fact, the slowdown due to change from CTC to Attention is $>10\times$ as compared to adding BiLSTM or TPS in the pipeline. In ViTSTR, the transition from tiny to small version requires an increase in embedding size and number of heads. No additional stage is necessary. The performance penalty to gain 2.3\% in accuracy is increase in number of parameters by 16.1M. From tiny to base, the performance penalty to gain 3.4\% in accuracy is additional 80.4M parameters. In both cases, the speed barely changed since we use the same parallel tensor dot product, softmax and addition operations in MLP and MSA layers of the transformer encoder. Only the tensor dimension is increased resulting to a minimal 0.2 to 0.3 msec/image slowdown in task completion.  Unlike in multi-stage STR, an additional module requires additional sequential layers of forward propagation which can not be parallelized resulting into a significant performance penalty.

\subsection{Failure Cases}
Table \ref{tab:failed_cases} shows sample failed predictions by ViTSTR-Small from each test dataset. The main causes of wrong prediction are confusion between similar symbols (e.g. 8 and B, J and I), scripted font (e.g. I in Inc), glare on a character, vertical text, heavily curved text image and partially occluded symbol. Note that in some of these cases, even a human reader can easily make a mistake. However, humans use semantics to resolve ambiguities. Semantics has been used in recent STR methods \cite{qiao2020seed,yu2020towards}.

\section{Conclusion}
ViTSTR is a simple single stage model architecture that emphasizes balance in accuracy, speed and computational requirements. With data augmentation targeted for STR, ViTSTR can significantly increase the accuracy especially for irregular datasets. When scaled up, ViTSTR stays at the frontiers to balance accuracy, speed and computational requirements.

\subsubsection*{Acknowledgements.}
This work was funded by the University of the Philippines ECWRG 2019-2020. GPU machines have been supported by CHED-PCARI AIRSCAN Project and Samsung R\&D PH. Special thanks to the people of Computer Networks Laboratory: Roel Ocampo, Vladimir Zurbano, Lope Beltran II, and John Robert Mendoza, who worked tirelessly during the pandemic to ensure that our network and servers are continuously running.

%
%
%
%
%
\bibliographystyle{splncs04}
\bibliography{egbib}
%






\end{document}